\documentclass[10pt,letterpaper]{article}

\usepackage{cogsci}

\cogscifinalcopy 

\usepackage{pslatex}
\usepackage{apacite}
\usepackage{float} 
\usepackage{graphicx}
\usepackage{xcolor}
\usepackage[tone]{tipa}
\usepackage{natbib}
\usepackage{multirow}
\usepackage{siunitx}
\usepackage{threeparttable}
\usepackage{url}
\usepackage[shortlabels]{enumitem}
\usepackage{microtype}
\usepackage{arydshln}
\usepackage[hidelinks]{hyperref}

\usepackage{color,soul}

\title{Evaluating computational models of infant phonetic learning across languages}

\author{{\large \bf Yevgen Matusevych (yevgen.matusevych@ed.ac.uk)}  \\
  School of Informatics, University of Edinburgh \\
  \AND {\large \bf Thomas Schatz (tschatz@umd.edu)} \\
  Department of Linguistics \& UMIACS, University of Maryland \\
  \AND {\large \bf Herman Kamper (kamperh@sun.ac.za)} \\
  E\&E Engineering, Stellenbosch University \\
  \AND {\large \bf Naomi H.\ Feldman (nhf@umd.edu)} \\
  Department of Linguistics \& UMIACS, University of Maryland \\
  \AND {\large \bf Sharon Goldwater (sgwater@inf.ed.ac.uk)} \\
  School of Informatics, University of Edinburgh }

\newcommand{\upd}[1]{#1}

\begin{document}

\maketitle

\begin{abstract}

In the first year of life, infants' speech perception becomes attuned to the sounds of their native language. Many accounts of this early phonetic learning exist, but computational models predicting the attunement patterns observed in infants from the speech input they hear have been lacking. A recent study presented the first such model, drawing on algorithms proposed for unsupervised learning from naturalistic speech, and tested it on a single phone contrast. Here we study five such algorithms, selected for their potential cognitive relevance. 
We simulate phonetic learning with each algorithm and perform tests on three phone contrasts from different languages, comparing the results to infants' discrimination patterns.
The five models display varying degrees of agreement with empirical observations, showing that our approach can help decide between candidate mechanisms for early phonetic learning, and providing insight into which aspects of the models are critical for capturing infants' perceptual development.

\end{abstract}

\section{Introduction}

Infants' speech perception changes in the first year of their life. For example, at the age of 6--8 months, English-learning and Japanese-learning infants are equally able to detect the difference between sounds \textipa{[\*r]} (as in \emph{rock}) and \textipa{[l]} (as in \emph{lock}), whereas by the age of 10--12 months, the two groups diverge, showing attunement to the phonetic contrasts present in their input language \citep{kuhl2006}. Similar results have been reported for many other languages \citep*[][etc.]{werker1984, bosch2003, tsao2006}. A number of theoretical accounts explaining such early phonetic learning have been proposed \citep[e.g.,][]{kuhl1995, best1994, werker2005}, but until recently no computational models could explain 
how the specific speech input to which infants are exposed leads to the observed changes in those infants' discrimination of phonetic contrasts. Models \emph{have} been evaluated on their ability to learn phonetic categories, rather than to predict patterns of discrimination (\citealp{vallabha2007}; \citealp*{mcmurray2009}), but none has yet succeeded in that task when learning from non-idealized natural speech input \citep{bion2013, antetomaso2017}. 

In a recent study, \citet{schatz2019} were the first to present such a computational model, which correctly predicted the documented cross-linguistic difference in infants' discrimination of \textipa{[\*r]} and \textipa{[l]} after learning from unsegmented speech. They explicitly simulated the learning process for Japanese and American English infants by (separately) training their model on naturalistic speech recordings either in Japanese or in American English. 
They then measured the trained models' ability to discriminate \textipa{[\*r]} and \textipa{[l]} with the machine ABX task, a flexible measure of discrimination that can be applied to model representations in essentially any format.
To obtain a model capable of handling realistic input, they selected an algorithm for unsupervised learning from naturalistic speech that had been proposed in the context of engineering applications.
The success of their model raises the question of whether other learning algorithms recently proposed in this context may lead to equally good, or even better, models of infants' early phonetic learning.

In this paper we apply the approach introduced in \cite{schatz2019} to test a variety of models on multiple data sets of infant phone discrimination. 
Doing so allows us 
to gain insight into the kinds of representations and learning mechanisms that infants are likely to employ.
We consider five different learning algorithms developed in the speech technology community, focusing on those that implement cognitively plausible learning mechanisms.
We test the models on three crosslinguistic phone discrimination tasks grounded in infant studies from different languages. 
Our two goals are (1) to test whether \citeauthor{schatz2019}'s result is specific to their particular model applied to a particular American English phonetic contrast, and (2) to identify which computational models can best explain the existing infant data.
We find that two models
show infant-like crosslinguistic discrimination patterns for American English \textipa{[\*r]}--\textipa{[l]} and another contrast in Mandarin Chinese, while three other models appear less successful as models of early phonetic learning. We perform additional analyses to assess whether the two most successful models---which substantially differ in their learning mechanisms and representation formats---also differ in what they learn. The results suggest that it should be possible to find an empirical test to decide between these two models. 

\section{Method}

In a series of simulations, we train computational models on unsegmented speech data from different languages. Each group of simulations focuses on one phonetic contrast (such as American English \textipa{[\*r]}--\textipa{[l]}) for which cross-linguistic phone discrimination data for infants exist. We test five different models on each contrast. For each model, we train two different versions: a `native' model, which simulates a learner of the language from which the contrast is drawn (American English, in this example), and a `non-native' model, which simulates a learner of another language that does not contain the relevant contrast (here, Japanese). Models are trained on corpora of natural speech, to simulate learning `in the wild'. We then test each model by simulating a phone discrimination task, using real examples from the language where the contrast exists. 
To show an infant-like pattern, the `native' trained version of the model should display better discrimination than the `non-native' trained version of the model. 

The training and test conditions for the three simulations are summarized in Table~\ref{tbl:exps}. The contrasts are chosen based on the strength of evidence regarding infants' behavior on this contrast and the availability of corresponding speech corpora for training computational models. 
{\bf Simulation 1} tests models learning American English and Japanese on the English \textipa{[\*r]}--\textipa{[l]} contrast, where English learners show better discrimination than Japanese learners by 12 months
\citep{kuhl2006,tsushima1994}.
{\bf Simulation 2} tests models learning Mandarin Chinese and American English on the Mandarin \textipa{[C]}--\textipa{[\texttctclig \super h]} contrast, where Mandarin learners show better discrimination than English learners by 12 months (\citealp{tsao2006}; \citealp*{kuhl2003}). {\bf Simulation 3} tests Catalan- and Spanish-learning models on the Catalan \textipa{[e]}--\textipa{[E]} contrast, where Catalan learners show better discrimination than Spanish learners by 8 months \citep*{bosch2003,albareda2011}. In the experiments with infants, each phonetic contrast was tested in a particular phonetic context (e.g., \textipa{[\*ra]}--\textipa{[la]}). To have sufficient test data for our models, we report the results averaged over all phonetic contexts instead. However, we also tested the models in the restricted contexts that were actually used in the experiments, and the trends were in the same direction.

\begin{table}
 \vspace{-1mm}
\centering
\caption{Training and test conditions.}
\label{tbl:exps}
\begin{tabular}{cp{1.6cm}p{2.2cm}p{1.7cm}}
\hline
\multicolumn{1}{l}{Sim.\ \#}  & Test \hspace{5mm} language & Training \hspace{5mm} language & Listener type             \\ \hline
\multirow{2}{*}{1}  & \multirow{2}{*}{EN}   & English (EN)  & Native  \\
                    &                       & Japanese (JA) & Non-native \\ \hline
\multirow{2}{*}{2}  & \multirow{2}{*}{ZH}   & Mandarin (ZH) & Native \\
                    &                       & English (EN)  & Non-native \\ \hline
\multirow{2}{*}{3}  & \multirow{2}{*}{CA}   & Catalan (CA)  & Native  \\
                    &                       & Spanish (ES)  & Non-native \\ \hline
\end{tabular}
 \vspace{-1mm}
\end{table}

\subsection{Simulating phone discrimination tasks}

To test a model's ability to discriminate a phonetic contrast, similar to the tests carried out with infants such as conditioned head turn \citep*{werker1997}, we use the machine ABX task \citep{schatz2013}.\footnote{\upd{\url{https://github.com/bootphon/ABXpy}}} In this task, A and X are two instances of the same phone (e.g., \textipa{[l]}), while B is a different phone (e.g., \textipa{[\*r]}). If A and X are closer to each other in a model's representation space than B and X, the model's prediction is correct, otherwise it is not.\footnote{Following earlier studies, we use Kullback--Leibler divergence to measure distances in the representations for one of the models (DPGMM, see Table~\ref{tbl:models}), and angular cosine distance for the other models. For models using frame-level representations (see Table~\ref{tbl:models}), we align the frames in each pair of phones using dynamic time warping \citep{vintsyuk1968}, and compute the distance as an average over the framewise distances.} 
A model is evaluated by considering the proportion of ABX triplets for which
it makes correct predictions: $0\%$ error rate corresponds to perfect discrimination, and $50\%$ to chance performance. Following \citet{schatz2019}, we sample ABX test triplets in such a way that all three phones---A, B, and X---appear in the same neighboring phonetic context and are pronounced by the same speaker. 

To test whether the difference between the ABX error rates in a given pair of simulated listeners (native vs.\ non-native) is significant, we fit mixed-effects regressions to the error rates of the two models in question. Each regression includes main effects of simulated listener type (native vs.\ non-native) and data subset (as shown in Table~\ref{tbl:traindata} below) and random intercepts, to account for the variation among data subsets, speakers and phonetic contexts. Significance for the effect of simulated listener type was then determined using two-tailed ANOVA tests (with Satterthwaite degrees of freedom approximation) on the predicted values of the regressions.

\subsection{Computational models}

We consider five models: the one used in \citet{schatz2019} and four neural network models inspired by existing work in unsupervised speech representation learning. These models show high performance in low-resource speech technology applications, making them a good starting point for modeling unsupervised infant learning. In high-level terms, the models differ along several dimensions, as summarized in Table~\ref{tbl:models}.
Three of the models learn representations at the level of speech frames (i.e., 25-ms-long chunks of speech commonly used in automatic speech recognition),
while two learn to encode word-sized units of variable length as vector representations of fixed length (somewhat similar to semantic word embeddings). \upd{Note, however, that at test time all models provide a way to compute distances between speech sequences (in this case, phones) of any duration.} Finally, three models are strictly unsupervised, while two others 
rely on top-down guidance from known wordforms---based on evidence that even 6--8-month-olds can segment and recognize some wordforms (\citealp{bortfeld2005,jus95a}; \citealp*{jus99b}). 
In all cases, we use existing implementations that were developed for processing speech with minimal supervision \citep{schatz2019,kamper2015,kamper2019}, \upd{adopt the previously used training options} and do not retune hyperparameters.

\noindent
\textbf{Dirichlet process Gaussian mixture model} \citep[DPGMM,][]{chen2015} \upd{with parallel MCMC sampling \citep{chang2013}} is the model used by \citet{schatz2019}. It is a probabilistic generative model with the number of Gaussian components (clusters) derived from the data. It learns in a fully unsupervised bottom-up manner, by soft-clustering individual speech frames. Using the learned mixture, each frame in the test data is represented as a posterior probability vector, consisting of the probabilities of this frame being generated by each component. We use the implementation based on \citet{chang2013}\footnotemark\ and parameter settings from \citet{schatz2019}: the model is initialized with $10$ clusters and is trained for $1501$ iterations.
\footnotetext{\url{http://people.csail.mit.edu/jchang7/code/subclusters/dpmm_subclusters_2014-08-06.zip}}

\noindent
\textbf {Autoencoder} \citep[AE,][]{kramer1991} is an unsupervised feedforward neural network. It learns by reconstructing (i.e., encoding and decoding) a given input, which results in the emergence of latent representations in its hidden layers. In this case, the model reconstructs individual speech frames one-by-one. We use a deep implementation with $8$ hidden layers ($7 \times 100$ and $1 \times 39$ units), as described in \citet{kamper2015},\footnote{\upd{\label{ae-cae}\url{https://github.com/kamperh/speech_correspondence}}} and pretrain it for $5$ epochs per layer plus $5$ epochs of final fine-tuning, \upd{without early stopping, using Adadelta optimization with adaptive learning rate \citep{zeiler2012} with decay $0.95$}. At test time, the 39-dimensional second-to-last hidden layer is used to encode individual frames from the test data into the model's representation space.

\noindent
\textbf {Correspondence autoencoder} \citep[CAE,][]{kamper2015} is a deep neural network similar to the AE, but it uses weak word-level supervision. Instead of trying to encode and reconstruct each input frame to itself, as is done in the AE, it is given pairs of spoken instances of the same word (aligned at the frame level), and tries to reconstruct each frame in one instance of a word from the aligned frame in the other instance---so the encoded representation must focus on linguistically meaningful information and abstract away from other variation between the aligned frames. Following \citet{kamper2015}, we initialize the CAE using the AE, and train the CAE with the same architecture as in the AE for $120$ epochs.\footnotemark[\value{footnote}]

\noindent
\textbf {Autoencoding recurrent neural network} \citep[AE-RNN,][]{chung2016}, or sequence-to-sequence autoencoder, is a type of AE in which both the encoder and the decoder are recurrent neural networks (RNNs). RNNs are commonly used in language modeling \citep{linzen2019}, as they can process an input sequence as a whole. In our case, the model is given a random word-sized chunk of speech, encodes it into a vector of fixed dimensionality, and then uses this vector to reconstruct the same chunk sequentially, frame-by-frame. We use the implementation by \citet{kamper2019},\footnotemark\ with $3$ hidden layers ($400$ gated recurrent units each) in both the decoder and the encoder, and an embedding dimensionality of $130$.
We train it for $15$ epochs \upd{without early stopping using
Adam optimization \citep{kingma2015} with a learning rate of $0.001$,}
and then use its encoder to convert each test chunk into a fixed-dimensional representation. 
\footnotetext{\url{https://github.com/kamperh/recipe_bucktsong_awe_py3}}

\noindent
\textbf {Correspondence-autoencoding recurrent neural network} \citep[CAE-RNN,][]{kamper2019} is similar to the AE-RNN, but instead of training on random chunks of speech, it is trained on pairs of instances of the same word---i.e., like the CAE, it also relies on weak top-down supervision. Again, following previous work \citep{kamper2019},\footnotemark[\value{footnote}] we use the AE-RNN to initialize the parameters of the CAE-RNN and then train it (with parameters analogous to those of the AE-RNN) for $3$ epochs.

\begin{table}[]
\vspace{-1mm}
\centering
\caption{Models of early phonetic learning used in the study.}
\label{tbl:models}
\begin{tabular}{lll}
\hline
Model   & Top-down guidance        & Representation type \\ \hline
DPGMM   & No              & Frames      \\
\hdashline
AE      & No              & Frames      \\
CAE     & Yes &    Frames   \\
\mbox{AE-RNN}  & No              & Word-sized   \\
\mbox{CAE-RNN} & Yes &    Word-sized             \\ \hline
\end{tabular}
\vspace{-1mm}
\end{table}

\subsection{Input to the models}

To prepare input to the models from unsegmented speech data, we follow a standard approach in speech processing, also adopted by \citet{schatz2019}: we discretize the speech data into 25-ms-long frames (sampled every 10 ms) and extract mel-frequency cepstral coefficients (MFCCs), together with their first and second time derivatives,
from each frame using Kaldi \citep{povey2011}. Representing speech using its auditory spectrum---as MFCCs do---is grounded in human auditory processing and is different from traditional accounts of phonetic learning,
which assume phonetic feature detectors \citep[see][for further discussion]{schatz2019}.

Additionally, we need to obtain speech alignments, i.e., labels that map chunks of speech to their corresponding phones or words. For training the models with top-down guidance, we obtain word-level alignments, and for testing all the models, we obtain phone-level alignments, using the Montreal Forced Aligner \citep{mcauliffe2017}. When running the alignments, we noticed that the Catalan dictionary did not always contain the correct entries for our target Catalan contrast, \textipa{[e]} vs.\ \textipa{[E]}, and we replaced such entries with standard Catalan pronunciations available in Wiktionary\footnote{\url{http://wiktionary.org}}. For Simulation~1, we obtained the existing alignments \citep{schatz2019} generated with Kaldi \citep{povey2011}.

In each case we train and test models on two different subsets of speech data per language, in order to ensure that the results for each model are robust across various data sets. Ideally, each subset should come from a different corpus, and the corpora should represent two different speech registers: spontaneous and read. In practice, our choices are limited to the available speech corpora, so that in Simulation 2 we use corpora of read speech only, and in Simulation 3 all our data come from the same bilingual corpus (Table~\ref{tbl:traindata}). \upd{To further reduce potential variability across corpora, we sample the audio signal in each corpus at $16$ kHz and balance the speakers' gender within each corpus sample.}

\begin{table}
\centering
\vspace{-2mm}
\caption{Corpus samples used in the simulations.}
\label{tbl:traindata}
\begin{threeparttable}
\begin{center}
A. Training data.
\end{center}
\begin{tabular}{p{0.4cm}p{1.1cm}p{1.6cm}p{1cm}>{\raggedleft\arraybackslash}p{1cm}>{\raggedleft\arraybackslash}p{0.9cm}}
\hline
\multicolumn{1}{p{0.4cm}}{Sim.\ \#} & Language         & Corpus      & Register           & Amount of data     & No.\ of spk.\\ \hline
\multirow{4}{*}{1}                & EN  & WSJ\tnote{1}         & Rd\tnote{7}               & 19:30 & 96 \\
                                  & JA & GlobalPhone\tnote{2} & Rd               & 19:33    & 96 \\
                                  \cline{2-6}
                                  & EN  & Buckeye\tnote{3}     & Sp\tnote{7}        & 9:13     & 20\\
                                  & JA & CSJ\tnote{4}         & Sp        & 9:11     & 20 \\ \hline
\multirow{4}{*}{2}                & ZH & AIShell\tnote{5}     & Rd               & 58:59    & 166 \\
                                  & EN  & WSJ         & Rd               & 58:49    & 166 \\
                                  \cline{2-6}
                                  & ZH & GlobalPhone & Rd               & 11:51    & 48 \\
                                  & EN  & WSJ         & Rd               & 11:49    & 48\\ \hline
\multirow{4}{*}{3}                & CA                   & Glissando\tnote{6} & Rd+Sp   & 7:41     & 26\\
                                  & ES                   & Glissando  & Rd+Sp   & 7:41    & 26\\
                                  \cline{2-6}
                                  & CA                   & Glissando   & Rd+Sp & 7:02     & 17\\
                                  & ES                   & Glissando   & Rd+Sp & 7:03    & 17 \\ \hline                                  
\end{tabular}
\begin{center}
B. Test data.
\end{center}
\begin{tabular}{p{0.4cm}p{1.1cm}p{1.6cm}p{1cm}>{\raggedleft\arraybackslash}p{1cm}>{\raggedleft\arraybackslash}p{0.9cm}}
\hline
\multirow{2}{*}{1}                & \multirow{2}{*}{EN}  & WSJ         & Rd               & 9:39     & 47\\
                                  &                           & Buckeye     & Sp        & 9:01     & 20 \\ \hline
\multirow{2}{*}{2}                & \multirow{2}{*}{ZH} & AIShell     & Rd               & 58:45    & 165 \\
                                  &                           & GlobalPhone & Rd               & 11:51    & 48 \\ \hline
\multirow{2}{*}{3}                                 & \multirow{2}{*}{CA}                   & Glissando   & Rd+Sp & 1:15     & 2 \\
                                 &                    & Glissando   & Sp & 2:19    & 11 \\ \hline
\end{tabular}
\begin{tablenotes}
           \item[1] {\footnotesize The Wall Street Journal CSR corpus \citep{paul1992}.}
           \item[2] {\footnotesize The multilingual text and speech database \citep{schultz2002}.}
           \item[3] {\footnotesize The Buckeye corpus of conversational speech \citep{pitt2005}.}
           \item[4] {\footnotesize Corpus of spontaneous Japanese \citep{maekawa2003}.}
           \item[5] {\footnotesize An open-source Mandarin speech corpus \citep{bu2017}.}
           \item[6] {\footnotesize A corpus for multidisciplinary prosodic studies in Spanish and Catalan \citep{garrido2013}.}
           \item[7] {\footnotesize Rd and Sp stand for read and spontaneous speech, respectively.}
     \end{tablenotes}
\vspace{-4mm}

\end{threeparttable}

\end{table}

\section{Results}

\subsection{Crosslinguistic patterns of ABX discrimination}

\begin{figure}[!h]
  \centering
  \includegraphics[width=\linewidth]{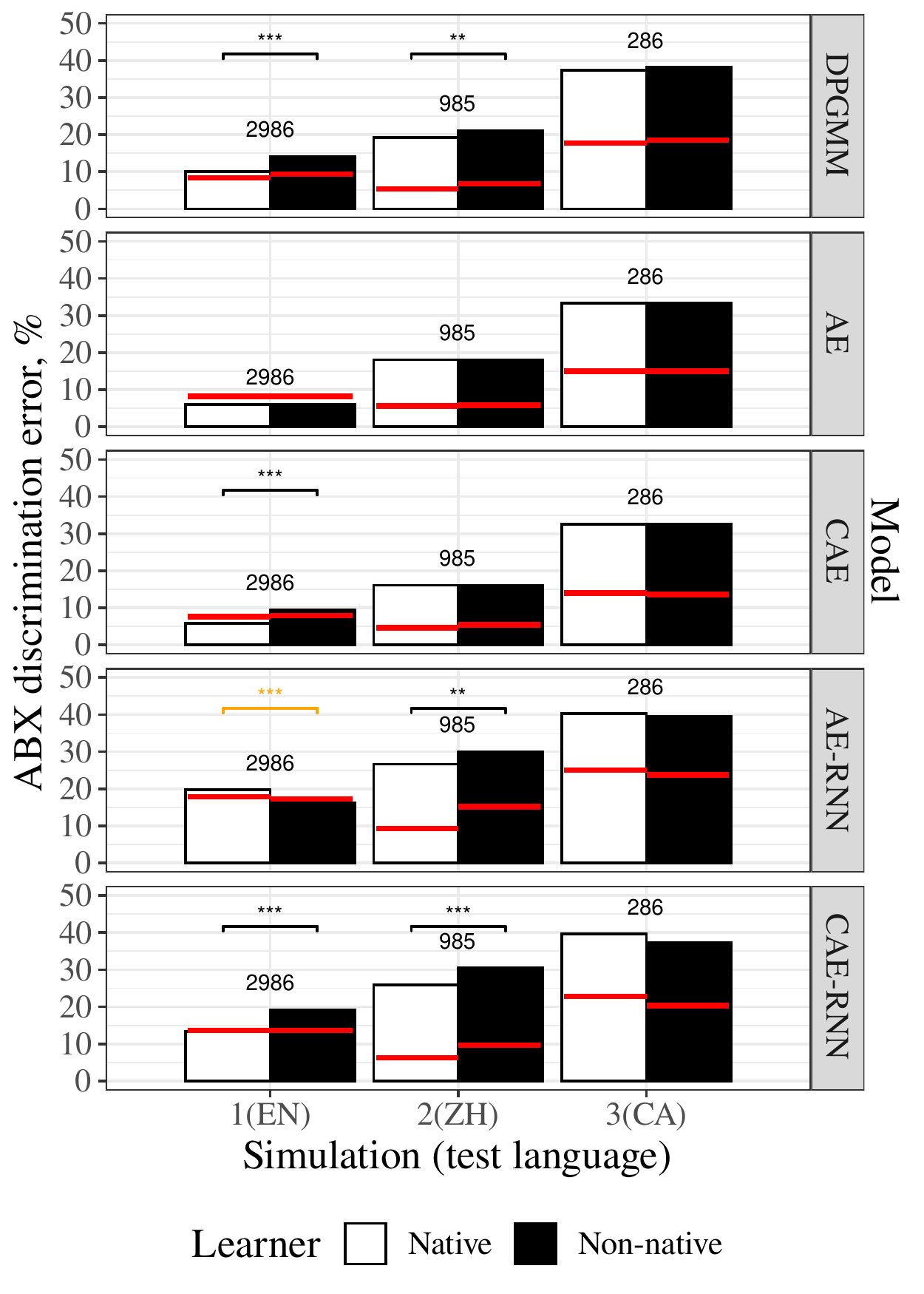}
   \vspace{-9mm}
  \caption{ABX error rates of the five native and non-native models in the three discrimination tasks (EN \textipa{[\*r]}--\textipa{[l]}, ZH \textipa{[C]}--\textipa{[\texttctclig \super h]} and CA \textipa{[e]}--\textipa{[E]}). To match the infant pattern of discrimination, the native model in each pair must show significantly lower error rates than the non-native model: out of $7$ patterns with a significant difference, $6$ are in the predicted direction (black brackets) and $1$ is in the wrong direction (orange bracket). The number of data pairs 
  in each test set is shown on top of each bar. Red lines indicate models' error rates averaged over all consonant (for EN and ZH) or all vowel (for CA) contrasts.}
    \vspace{-4mm}
  \label{fig:results}
\end{figure}

The ABX error rates of each model across languages are shown in Figure~\ref{fig:results}, together with the average performance of each model across all consonant (for English and Mandarin Chinese) or vowel (for Catalan) contrasts (red lines in the figure). In this figure, results are averaged over multiple ABX triplets, speakers, neighboring phonetic contexts, and subsets of the corpus, although the significance values take into account all of these variables 
(as previously discussed).  The reported patterns are consistent over the two corpus subsets in all cases, except the AE-RNN on the Mandarin contrast (where the difference between the `native' and the `non-native' model does not reach significance when trained on the smaller subset). In what follows, we compare the performance of each simulated `native' listener to its corresponding simulated `non-native' listener. Note that comparing the results \textit{across} the simulations is not meaningful, as the amount of training data and number of speakers differed depending on the simulation.

In Simulation 1, which tests the models on the English \textipa{[\*r]}--\textipa{[l]} contrast, three out of five models (DPGMM, CAE and CAE-RNN) correctly predict the discrimination pattern observed in infants: the error rate of the simulated native listener is significantly lower compared to the simulated non-native listener. Two models do not predict this pattern correctly: the AE shows no significant difference between the two types of simulated learners, and the AE-RNN predicts a significant difference in the wrong direction (i.e., lower error rate in the non-native listener). Across all models (native and non-native), the discrimination error on this contrast ranges from $5.8$--$19.7\%$, comparable to the average error on all English consonant contrasts (red lines, $7.6$--$17.8\%$). In other words, the \textipa{[\*r]}--\textipa{[l]} contrast is easy to discriminate for all models. 

In Simulation 2, with the Mandarin \textipa{[C]}--\textipa{[\texttctclig \super h]} contrast, a different set of three models (DPGMM, \textit{AE-RNN} and CAE-RNN) correctly predict the infants' discrimination pattern, and two models (AE and CAE) predict no significant difference between the simulated native and the non-native listener. 
Over all models, the discrimination error on this contrast ranges from $16.0$--$30.5\%$---noticeably higher than the average error over all Mandarin consonant contrasts (red lines, $4.6$--$9.6\%$).
In other words, the [\textipa{C}]--[\textipa{\texttctclig \super h}] discrimination is difficult 
for the models. This may be due to the kinds of phones in this contrast: one of them, \textipa{[\texttctclig \super h]}, is a combination of a short \textipa{[t]} followed by a `breathy' version of \textipa{[C]}, the other phone in the contrast (somewhat similar to the distinction between the first phones in \textit{cheap} vs.\ \textit{sheep}). As a result, one of the phones, \textipa{[C]}, is almost a `subchunk' of the other phone, \textipa{[\texttctclig \super h]}, a distinction potentially difficult to learn for our models. Nevertheless, three `native' models show lower error rate than the corresponding `non-native' models.

In Simulation 3, no model predicts a significant difference for the Catalan \textipa{[e]}--\textipa{[E]} contrast.
In general, this contrast is more difficult for all models than an average Catalan vowel contrast, with error rates ranging from $32.5$--$40.3\%$ (vs.\ the average of $14.0$--$23.7\%$). Further analysis (not shown) revealed that this may be because some \textipa{[e]} and \textipa{[E]} vowels in the test data had very short duration (unlike the lab stimuli used with infants). Filtering out very short ($<80$~ms) phones from the test data reduced the overall error rates, but still yielded similar performance between the `native' and `non-native' models. \upd{Note that the models' average error rates on Catalan are generally high, suggesting that the models could benefit from additional training data. At the same time, speaker idiosyncrasies in the test data are unlikely to affect the results, as we observe no meaningful differences in the discrimination error across the two test samples (consisting of data from $2$ vs.\ $11$ speakers).} 
Thus, all models struggle to discriminate the Catalan \textipa{[e]}--\textipa{[E]} contrast, as well as to reproduce empirically observed cross-linguistic differences in its discrimination. 

While no model captures all three crosslinguistic discrimination patterns, the DPGMM and the CAE-RNN correctly predict two patterns out of three. The CAE and the AE-RNN only predict one pattern each, while the AE makes no correct predictions. The two models that perform best use very different learning algorithms and representation formats, effectively presenting two alternative hypotheses about early phonetic learning, yet they make identical predictions regarding the crosslinguistic discrimination of three phone contrasts. This raises the question of whether these two models, despite their differences, may behave similarly on discrimination tasks in general. We address this question in the next section. 

\subsection{Comparing predictions of the two models}

In this section, we identify phone contrasts for which the two models---DPGMM and CAE-RNN---make different predictions in the discrimination task. For simplicity, we only focus on the `native' listeners---i.e., trained and tested on the same language. We use $5$ simulated listeners trained on different languages: English and Japanese from Simulation~1, Mandarin from Simulation~2,
Catalan and Spanish from Simulation~3. We test each model on a variety of contrasts from its training language. Among hundreds of phone contrasts that could be tested, many of them are trivial to discriminate (e.g., a vowel vs.\ a consonant), and we only consider those where the two phones differ on one distinctive phonological feature: e.g., Japanese \textipa{[i]} and \textipa{[i:]} differ in \textit{length} (i.e., duration), and Mandarin \textipa{[p]} and \textipa{[p\super h]} differ in \textit{aspiration} (i.e., `breathiness').

\begin{table}
\vspace{-2mm}
\centering
\caption{Phone contrasts with the largest difference in ABX discrimination between the two models in terms of standard scores (or $z$-scores: how many standard deviations the error rate of a particular model is higher or lower than the mean error rate of that model). Positive difference indicates that the contrast is easier for the CAE-RNN than for the DPGMM.} \vspace{2mm}
\label{tbl:difficulty}
\begin{tabular}{p{0.9cm}p{1.1cm} *{3}{S[detect-weight,table-format=1.1, retain-explicit-plus]}}
\hline
  &   & \multicolumn{3}{c}{Standard score} \\
\cline{3-5}
\multicolumn{1}{p{0.9cm}}{Contrast} &
\multicolumn{1}{p{1.1cm}}{Language} &
\multicolumn{1}{p{1.1cm}}{DPGMM} & 
\multicolumn{1}{p{1.6cm}}{CAE-RNN} &
\multicolumn{1}{p{1.6cm}}{Difference} \\
\hline
\textipa{d}--\textipa{z} &   JA & -0.2  & -1.0  & +0.8    \\
\textipa{i}--\textipa{i:} &   JA & -0.4  & -1.2 & +0.8    \\
\textipa{a}--\textipa{a:} &   JA  & -0.4  &  -1.1 & +0.7   \\
\textipa{p\super h}--\textipa{t\super h} &   ZH  & -0.2  & -0.8 & +0.6   \\
\textipa{r}--\textipa{z} &   JA  & -0.8  & -1.4 & +0.6   \\
\textipa{C}--\textipa{C:} &   JA  & +0.2  &  -0.3  & +0.5\\

\hline
\textipa{h}--\textipa{T} &   EN  & -1.4  & 0.0    & -1.4    \\
\textipa{g}--\textipa{\ng} &   EN  & -0.9  & +0.4 &   -1.3    \\
\textipa{u}--\textipa{U} &   EN  & -0.3 & +0.9  & -1.2   \\
\textipa{f}--\textipa{x}    &      ES     & -0.4  & +0.6    & -1.0 \\
\textipa{t}--\textipa{T} &   ES  & -0.7 & +0.3    & -1.0\\
\textipa{t}--\textipa{t:} & JA  & -0.7  &  +0.3 & -1.0\\
\hline
\end{tabular}
\vspace{-3mm}
\end{table}

Table~\ref{tbl:difficulty} shows the contrasts for which the discrimination performance of the DPGMM and the CAE-RNN differs the most. 
As the average error rates (and their distributions) vary across the two models, we report differences 
in terms of standard scores. One pattern we see is that the easiest contrasts for the CAE-RNN relative to the DPGMM (upper part of Table~\ref{tbl:difficulty}) are those where the two phones are continuous and
differ only in their length: Japanese \textipa{[i]}--\textipa{[i:]} and \textipa{[a]}--\textipa{[a:]}, Mandarin \textipa{[C]}--\textipa{[C:]}. The DPGMM performs best relative to the CAE-RNN (lower part of Table~\ref{tbl:difficulty}) on such contrasts as \textipa{[t]}--\textipa{[t:]}
and \textipa{[u]}--\textipa{[U]}, where one phone (here, \textipa{[U]} and \textipa{[t]}) is similar to a `subchunk' of the other phone (\textipa{[u]} and \textipa{[t:]}), so that the difference between the two is only observable within a short time window: e.g., \textipa{[t:]} is a silence followed by \textipa{[t]} \citep[and see][regarding the vowel contrast]{hillenbrand1995}. This may be because the CAE-RNN `compresses' the time dimension into a fixed-dimensional representation, losing the information about individual speech frames. It is less clear why contrasts involving \textipa{[T]} are difficult for the CAE-RNN, though \citet{jongman1989} shows that listeners need to hear the full duration of \textipa{[T]} to successfully discriminate it from similar sounds.

We also looked at the models' error rates on \textit{types} of contrasts (e.g., all contrasts where the two phones differ on a particular feature), to see whether the discrimination along a particular phonological feature (e.g., aspiration) is easier for one model than for the other. The most salient pattern is that
length contrasts (e.g., short \textipa{[a]} vs.\ long \textipa{[a:]}, cf.\ Table~\ref{tbl:difficulty}) are on average easier to discriminate for the CAE-RNN (compared to other types of contrasts) than for the DPGMM, in line with our analysis for the individual contrasts.

\section{Conclusion}

Using computational modeling on realistic input, we compared possible mechanisms for early phonetic learning in their ability to predict the changes in discrimination empirically observed in infants. We tested five models on three phone contrasts from different languages, using a phone discrimination task. In our study, \citeauthor{schatz2019}'s (\citeyear{schatz2019}) DPGMM model showed the infant-like crosslinguistic pattern of discrimination both for English \textipa{[\*r]}--\textipa{[l]} and for Mandarin \textipa{[C]}--\textipa{[\texttctclig \super h]}, demonstrating that the earlier result was not specific to a particular English contrast. Moreover, a second model, the CAE-RNN, also made correct predictions on the same two contrasts.
Although no model predicted the correct pattern for the Catalan vowel contrast \textipa{[e]}--\textipa{[E]}, the fact that two of them make correct predictions on the other two contrasts is promising. This result supports the idea that models learning representations directly from unsegmented natural speech can correctly predict some of the infant phone discrimination data. Based on the results of this study, the DPGMM and the CAE-RNN show some promise as models of early phonetic learning, although their failure to capture the effect on the Catalan contrast warrants further investigation to determine whether the failure is a problem with the models or with the amount and/or quality of the training and the test data. 

Here we are making predictions about infants' phone discrimination. Adults show many cross-linguistic discrimination differences as well, and one question for future research is how the presented models of infant speech perception relate to predictions one might make about adult perception. There are likely to be additional learning mechanisms later in childhood beyond those we have modeled here, but what changes over development, and how, remains an open question.

The DPGMM and the CAE-RNN both aim to model primarily unsupervised infant phonetic learning, although they
make very different assumptions about the learning mechanism and the types of representations: the DPGMM is an unsupervised bottom-up model that learns by clustering frame-level representations, while the CAE-RNN encodes chunks of speech holistically, learns by sequentially reconstructing an acoustic word, and uses weak top-down guidance from the word level \citep[but see][for a fully unsupervised alternative]{kamper2019}.
Importantly, our analysis shows that the two models do not always make identical predictions about discriminability: for example, vowel length contrasts are generally easier to discriminate for the CAE-RNN than for the DPGMM. Such differences between the models allow us to make predictions---both for native and non-native listeners---which can then be tested in experimental studies with infants to differentiate between the models.

\section{Acknowledgements}

\upd{This work is based on research supported in part by an NSF award BCS-1734245 and an ESRC-SBE award ES/R006660/1. We thank the anonymous reviewers, the members of Maryland Phonology Circle, and Kate McCurdy for helpful feedback.}

\setlength{\bibleftmargin}{.125in}
\setlength{\bibindent}{-\bibleftmargin}

\bibliographystyle{apacite}
\bibliography{references}

\end{document}